\pdfoutput=1
\documentclass{article}

\PassOptionsToPackage{numbers}{natbib}
\usepackage[preprint]{arxiv}

\usepackage[utf8]{inputenc} 
\usepackage[T1]{fontenc}    
\usepackage{hyperref}       
\usepackage{url}            
\usepackage{booktabs}       
\usepackage{amsfonts}       
\usepackage{nicefrac}       
\usepackage{microtype}      
\usepackage{comment}        
\usepackage{amsmath}
\usepackage{multirow}
\usepackage{bm}
\usepackage{amsthm}
\usepackage{hhline}
\usepackage{amssymb}
\usepackage{makecell}
\usepackage{mathtools}
\usepackage{enumitem}
\usepackage{color}
\usepackage{xcolor}
\usepackage{MnSymbol}
\usepackage{makecell}
\usepackage{graphicx}
\usepackage{arydshln}
\usepackage{booktabs}
\usepackage{framed}
\usepackage[font=small]{caption}
\usepackage[scientific-notation=true]{siunitx}
\usepackage[most]{tcolorbox}
\usepackage{wrapfig}
\usepackage{lipsum}
\usepackage{sidecap}
\usepackage{pifont}
\usepackage{fixltx2e}
\usepackage[flushleft]{threeparttable}
\usepackage{diagbox}

\newtcolorbox[auto counter, number within=section]{definitionbox}[2][]{%
  colframe=blue!5!white,, 
  colback=blue!5!white,  
  before upper={#2},
  #1
}

\title{\sc A Vision for Multisensory Intelligence:\\Sensing, Science, and Synergy}

\author{%
    Paul Pu Liang\\
    Multisensory Intelligence Group\\
    MIT Media Lab and MIT EECS\\
    \texttt{\href{mailto:ppliang@mit.edu}{ppliang@mit.edu}}
}

\begin{document}

\maketitle

\vspace{-6mm}
\begin{abstract}
Our experience of the world is multisensory, spanning a synthesis of language, sight, sound, touch, taste, and smell. Yet, artificial intelligence has primarily advanced in digital modalities like text, vision, and audio. This paper outlines a research vision for multisensory artificial intelligence over the next decade. This new set of technologies can change how humans and AI experience and interact with one another, by connecting AI to the human senses and a rich spectrum of signals from physiological and tactile cues on the body, to physical and social signals in homes, cities, and the environment. We outline how this field must advance through three interrelated themes of \textit{sensing}, \textit{science}, and \textit{synergy}. Firstly, research in sensing should extend how AI captures the world in unconventional ways beyond the digital medium. Secondly, developing a principled science for quantifying multimodal heterogeneity and interactions, developing unified modeling architectures and representations, and understanding cross-modal transfer. Finally, we present new technical challenges to learn synergy between modalities and between humans and AI, covering multisensory integration, alignment, reasoning, generation, generalization, and experience. Accompanying this vision paper are a series of projects, resources, and demos of latest advances from the Multisensory Intelligence group at the MIT Media Lab, see \url{https://mit-mi.github.io/}.
\end{abstract}

\vspace{-3mm}
\section{Introduction}
\vspace{-1mm}

Over the past decade, multimodal interaction has become a central theme in AI and HCI research~\citep{oviatt1999ten,stivers2005introduction,turk2014multimodal}. Early work explored how systems could understand and respond across multiple modalities, such as language, vision, and audio, through channels like speech interfaces~\citep{oviatt2007multimodal}, embodied agents~\citep{xie2024large}, and affective computing~\citep{picard2000affective}. These historical and recent advances asked fundamental questions about how humans express intent and emotion across modalities~\citep{poria2017review}, how to align heterogeneous signals~\citep{rasenberg2020alignment}, and how to design interfaces that support learning and interaction~\citep{karpov2018multimodal}. Fueled by innovations in large-scale multimodal datasets and deep learning~\citep{liang2024foundations}, it has now culminated in today's large-scale multimodal foundation models that integrate text, images, and audio with unified architectures~\citep{hurst2024gpt,team2023gemini} and enabling unprecedented generalization and transfer~\citep{radford2021learning,reed2022generalist}.

As these capabilities mature, a new frontier has emerged: multisensory intelligence. While multimodal AI has focused primarily on digital modalities (language, vision, audio), multisensory intelligence seeks to go deeper by connecting AI to the human senses and the physical world~\citep{cangelosi2015embodied,sharma2002toward}, spanning the rich spectrum of signals we and our environments produce~\citep{ghazal2021iot,krishnamurthi2020overview}, from physiological and tactile cues on the body~\citep{dahiya2009tactile,shu2018review}, to environmental and social signals in homes, cities, and ecosystems~\citep{adib2015smart,campbell2006people,tapia2004activity}. Furthermore, contrary to traditional models focusing on prediction and reasoning, multisensory intelligence envisions an end-to-end pipeline from sensor design, data collection and processing, AI modeling, and interactive deployments to improve human experiences in physical and social contexts~\citep{mathur2024advancing,pfeifer2004embodied,savva2019habitat}. We believe in the immense impact of multisensory intelligence to enhance how humans and AI experience and interact with one another, leading to improved productivity, creativity, and wellbeing for the world.

The objective of this paper is to articulate a vision for multisensory intelligence research in the next decade through three interrelated themes. Firstly, \textit{sensing} - the process of perceiving and capturing signals from the world (inspired by and extending the human senses) and transforming them into structured representations that support learning, reasoning, and decision-making (section~\ref{sec:sensing}). Secondly, \textit{science}, where section~\ref{sec:science} discusses current and future directions on developing a systematic understanding and learning of generalizable principles from multisensory data: some of the most important questions include developing a principled science for quantifying multimodal interactions, developing unified modeling architectures and representations, and understanding cross-modal transfer. Finally, section~\ref{sec:synergy} presents new technical challenges aiming to learn \textit{synergy}, the emergent integration of multiple modalities to produce intelligent capabilities greater than the sum of their parts. These challenges are multifaceted, covering multisensory integration, alignment, reasoning, generation, generalization, and experience. While the first few challenges seek to advance multisensory AI, the final challenges aim for synergistic interaction and co-intelligence between humans and multisensory AI.

We believe this survey paper will be useful for new researchers interested in AI, HCI, multimodal sensing, and their intersections, but we hope experienced practitioners will find it valuable as well. Accompanying this written medium is a series of projects, resources, and demos of latest advances from the Multisensory Intelligence group at the MIT Media Lab, see \url{https://mit-mi.github.io/}, and we encourage community involvement and expansion of these visions.

\vspace{-1mm}
\section{Novel Sensory Modalities}
\label{sec:sensing}
\vspace{-1mm}

\begin{figure}
    \centering
    \vspace{-4mm}
    \includegraphics[width=\linewidth]{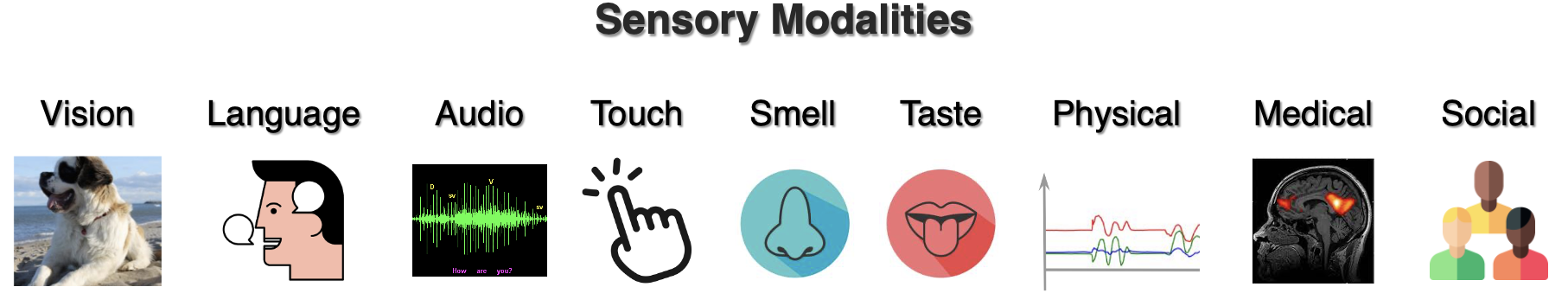}
    \vspace{-2mm}
    \caption{\textbf{Theme 1: Sensing} - the process of perceiving and capturing signals from the world (inspired by and extending the human senses) and transforming them into structured representations that support learning, reasoning, and decision-making. Section~\ref{sec:sensing} covers novel ways of sensing the world and the new challenges these sensing modalities introduce to AI research.}
    \label{fig:sensing}
\end{figure}

In this section, we highlight the potential for AI to revolutionize the sensing, perception, reasoning, and decision-making over novel sensory modalities. Our discussion covers the `traditional' language, vision, and audio modalities commonly seen in digital mediums, and will also highlight the next generation of modalities for AI.

\textbf{Vision} was the first modality to experience the deep learning revolution. Starting with the release of the Imagenet visual recognition challenges~\citep{deng2009imagenet} and early convolutional neural networks~\citep{lecun2002gradient}, recent years has seen a huge body of work for 2D vision in images~\citep{dosovitskiy2020image} and videos~\citep{madan2024foundation}, and advances in 3D~\citep{wang2025vggt} and 4D~\citep{zhou2025page} foundation models. Beyond pixels, other visual modalities include unconventional ways of seeing the world with wireless signals~\citep{adib2013see}, thermal imaging~\citep{ring2012infrared}, depth cameras~\citep{horaud2016overview}, and SPAD cameras~\citep{guerrieri2010two}, enabling seeing through walls~\citep{adib2013see} and around corners~\citep{kirmani2009looking,velten2012recovering}. Today, we have large-scale foundation models capable of solving many visual tasks at the same time~\citep{yuan2021florence}, with broader impacts in medical vision~\citep{esteva2021deep}, understanding human gestures and social interactions~\citep{robie2017machine}, in engineering systems and manufacturing~\citep{zhou2022computer}, and imaging from the depths of oceans~\citep{jang2019underwater} to outer space~\citep{akiyama2019first}.

\textbf{Language} has arguably been the primary driver for recent foundation modeling research, through Transformers~\citep{vaswani2017attention} and large language models~\citep{hurst2024gpt,team2023gemini}. Language tasks have also evolved to better reflect modes of human communication -- from typed text on computers and smartphones to spoken language~\citep{jurafsky2000speech}. Spoken language, in particular, is a multimodal problem involving fusing verbal language with nonverbal visual and vocal expressions~\citep{tsai2019multimodal}. To improve the accessibility of language technologies, research has expanded to include different languages, including low-resource languages~\citep{hedderich2021survey}, sign language~\citep{chua2025emosign}, and performative language~\citep{li2025mimeqa}.

\textbf{Audio} is a key component of human communication, carrying rich layers of information beyond the spoken word~\citep{jurafsky2000speech}. It encodes vocal expressions, prosody, rhythm, and paralinguistic cues that reveal affect, intent, and social dynamics~\citep{russell2003facial}. Beyond communication, audio is also central to creative and cultural expression through music, performance, and its interplay with visual and physical media such as art, fashion, and movement. Together, language, vision, and audio form the basis for multimodal human communication~\citep{ong2025human}, spanning spoken language, facial expressions and gestures, body language, vocal expressions, and prosody~\citep{picard2000affective}. Common tasks include predicting sentiment~\citep{soleymani2017survey}, emotions~\citep{zadeh2018multimodal}, humor~\cite{hasan2019ur}, and sarcasm~\citep{castro2019towards} from multimodal videos of social interactions.

\textbf{Touch.} \ The sense of touch is critical to how human perceive and interact with the world. There are exciting directions in developing tactile sensors, such as gloves~\citep{murphy2025fits,song2025opentouch,sundaram2019learning}, gelsight~\citep{yuan2017gelsight}, and digit~\citep{lambeta2020digit} sensors to capture the sense of touch at scale. Many of today's tangible and embodied AI systems are often equipped with multiple sensors to aid in robust decision-making for real-world physical tasks such as grasping, cleaning, and delivery~\citep{ishii1997tangible,lee2019making}. These multisensor robots have also been successfully applied in haptics~\citep{pai2005multisensory,seminara2019active} and surgical applications~\citep{abiri2019multi,bethea2004application}.

\textbf{Smell and taste} are understudied modalities that have the potential to revolutionize human perception and interaction with the world. Smell sensing can be performed using gas sensors to detect volatile organic compounds (VOCs)~\citep{sung2024data,yan2015electronic} and larger devices (e.g., GC-MS)~\citep{brattoli2011odour,leePrincipalOdorMap2023} to track fine-grained chemical compositions. This has enabled the curation of large-scale datasets to classify real-world substances based on gas sensor readings~\citep{feng2025smellnet} or molecular properties~\citep{snitzPredictingOdorPerceptual2013}. In addition to smell sensing, recent work has developed new methods for smell releasing and transmission~\citep{brooks2023smell} based on wearable devices~\citep{amoresEssenceOlfactoryInterfaces2017,liAromaBiteAugmentingFlavor2025, mayumiBubblEatDesigningBubbleBased2025,wangOnFaceOlfactoryInterfaces2020}. Finally, new senses include sensing and transmitting taste~\citep{brooks2023taste}, temperature regulation~\citep{brooks2020trigeminal}, and mouth haptics (i.e., mouthfeel)~\citep{shen2022mouth} are poised to inspire new research and to make AI closer to enhancing real-world human experiences.

\textbf{Physical sensing.} \ The physical world is witnessing an unprecedented surge in Internet of Things embedded with sensors, software, and communication technologies that can safely and privately analyze the human and physical world~\citep{li2015internet,rose2015internet}, including human physical activities~\citep{qi2015survey,yuehong2016internet}, vision, depth, and lidar for transportation~\citep{javaid2018smart,khayyam2020artificial}; and sensing for health~\citep{ahamed2018applying,kulkarni2014healthcare}. In the physical sciences, deepening our knowledge of nature and environments can bring about impactful changes in scientific discovery, sustainability, and conservation. This requires processing modalities such as chemical molecules~\citep{su2022molecular}, protein structures~\citep{zhang2019multimodal}, satellite images~\cite{cheng2017remote}, remote sensing~\citep{li2022deep}, wildlife movement~\citep{lopes2008development}, scientific diagrams and texts~\cite{lu2022learn}, and various physical sensors~\citep{mo2023multiiot}. There is also exciting work in AI for environmental~\citep{tuia2022perspectives}, climate~\citep{rolnick2022tackling}, ocean~\citep{jang2019underwater}, and space~\citep{furano2020ai} sensors, which can inspire the next generation of AI for scientific impact. Finally, future AI systems can also assist in scientific discover through hypothesis generation, lab-in-the-loop experimentation, and automated data analysis~\cite{angers2025roboculture,buehler2024accelerating, callahan2024open}.

\textbf{Medical sensing.} \ AI can help integrate complementary medical signals from lab tests, imaging reports, patient-doctor conversations, and multi-omics data to assist doctors in clinical practice~\cite{acosta2022multimodal,dai2025qoq,dai2025climb}. Multimodal physiological signals recorded regularly from smartphones and wearable devices can also provide non-invasive health monitoring~\cite{de2015multimodal,garcia2018mental}. Public datasets include MIMIC~\citep{MIMIC} with patient tabular data, medical reports, and medical sensor readings, question answering on pathology~\cite{he2020pathvqa} and radiology~\cite{lau2018dataset} images, and multi-omics data integration~\cite{tran2021openomics}.

\textbf{Social sensing.} \ Finally, AI can be used to sense the social world. In addition to verbal and nonverbal communicative modalities, these signals can come from smartphones and wearable devices~\citep{de2015multimodal,garcia2018mental,liang2021learning}, ubiquitous environments~\citep{mohr2017personal}, and social networks~\citep{pentland2014social}. Affective computing studies the perception of human affective states such as emotions, sentiment, and personalities from multimodal human communication: spoken language, facial expressions and gestures, body language, vocal expressions, and prosody~\citep{picard2000affective}. Some commonly studied tasks involve predicting sentiment~\citep{soleymani2017survey}, emotions~\citep{zadeh2018multimodal}, humor~\cite{hasan2019ur}, and sarcasm~\citep{castro2019towards} from multimodal videos of social interactions.

\vspace{-1mm}
\section{Scientific Foundations of Multisensory AI}
\label{sec:science}
\vspace{-1mm}

In this section, we discuss the new foundational research questions that emerge as we develop the next generation of multisensory intelligence. We first discuss the scientific questions of quantifying and modeling extremely heterogeneous sensing modalities, followed by how modalities interconnect to bring new task-relevant information. On the modeling side, we discuss scientific questions on learning more unified representations, better optimizing multimodal models, and encouraging cross-modal information transfer.

\subsection{Sensory heterogeneity}

The principle of heterogeneity reflects the observation that the information present in different modalities will often show diverse qualities, structures, and representations. There are several sources of heterogeneity - in the basic modality \textit{elements} (unit of data which in which the modality is represented)~\citep{barthes1977image}, the \textit{distribution}, frequency, and likelihood of elements in modalities, the \textit{structure} exhibited in the way individual elements are composed to form entire modalities~\citep{bronstein2021geometric} (e.g., discrete vs continuous, spatial vs temporal vs hierarchical), the \textit{information} content present in each modality~\citep{shannon1948mathematical}, the natural types of \textit{noise} and \textit{imperfections}~\citep{liang2021multibench}, and the types of \textit{tasks} that modality is most relevant for~\citep{zheng2024heterogeneous}. Studying and modeling heterogeneity is especially relevant today in the era of foundation models, as there is an aim to curate increasingly diverse multimodal pre-training datasets to enable transfer to heterogeneous downstream tasks~\citep{awadalla2023openflamingo,gadre2023datacomp}. There are key scientific questions in how to quantify types of heterogeneity, how to structure modeling architectures and training objectives depending on heterogeneous structure, whether we need to revisit inductive biases as large-scale datasets are increasingly saturated, and whether the future will lie in more modular, heterogeneity-aware versus unified, heterogeneity-agnostic models. Finally, data heterogeneity is a key driver in many fields beyond multimodal learning, including understanding distribution shifts, transfer learning, multitask learning, distributed and federated learning, even adversarial examples and anomaly detection~\citep{zamir2018taskonomy}.

\begin{figure}
    \centering
    \vspace{-4mm}
    \includegraphics[width=0.9\linewidth]{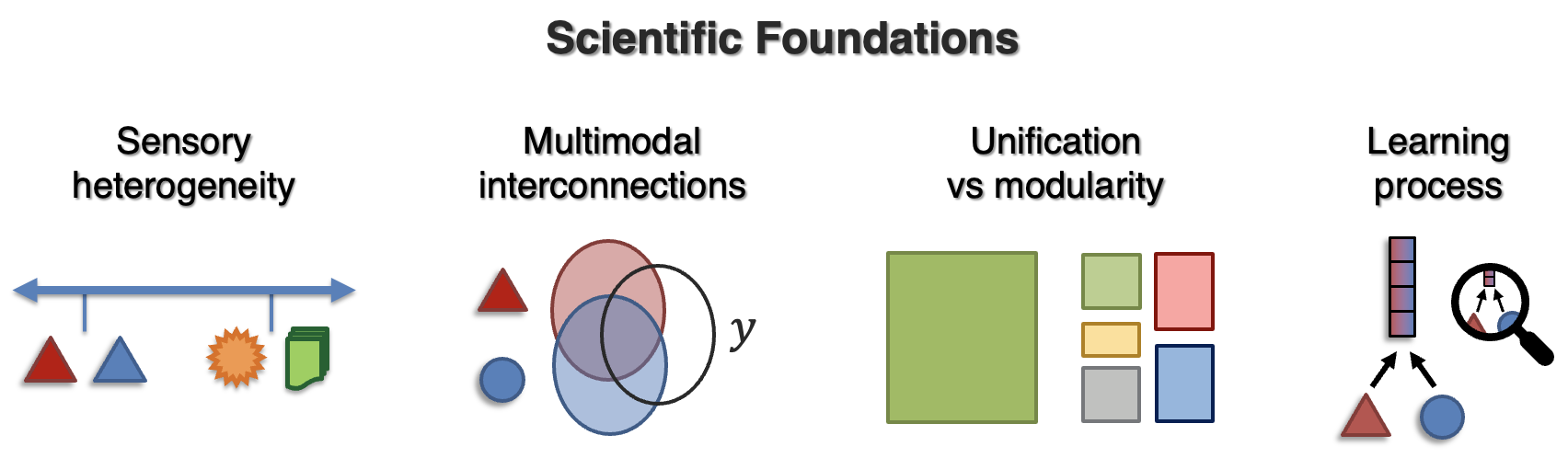}
    \vspace{-1mm}
    \caption{\textbf{Theme 2: Science} seeks to develop a systematic understanding and learning of generalizable principles from multisensory data: some of the most important scientific questions include (1) quantifying \textit{sensory heterogeneity} and its impact on modeling and training, (2) the types of \textit{multimodal connections and interactions} that give rise to new information during fusion, (3) characterizing the \textit{unified structures and representations} that scale across modalities and enable cross-modal generalization, and (4) advancing the \textit{learning dynamics and optimization landscapes} that shape practical multimodal learning.}
    \label{fig:science}
\end{figure}

\subsection{Multimodal interconnections}

Although modalities are heterogeneous, they are often connected and interact. Modality connections describe information \textit{sharing} across modalities, in contrast to \textit{unique} information that exists solely in a single modality~\citep{williams2010nonnegative}. Shared connections can be identified from distributional patterns in multimodal data~\citep{tian2020makes,turney2005corpus} or based on our domain knowledge~\citep{barthes1977image,marsh2003taxonomy}. While connections exist within multimodal data itself, modality interactions study how modalities combine to give rise to new information when integrated together for task \textit{inference}. There are several taxonomies useful for studying types of interactions: whether the integration is redundant (information shared between modalities, such as smiling while telling an overtly humorous joke); uniqueness (the information present in only one, such as each medical sensor designed to provide new information); and synergy (the emergence of new information using both modalities, such as conveying sarcasm through disagreeing verbal and nonverbal cues)~\citep{liang2023quantifying,williams2010nonnegative}; this broad taxonomy also permits more fine-grained analysis of interactions (e.g., whether responses are equivalent, enhanced, modulated, or emerges with multimodal inputs~\cite{partan1999communication}). Finally, interactions can also be studied based on the mechanics involved when integrated for a task (e.g., additive, non-additive, and non-linear forms~\citep{jayakumar2020multiplicative}, logical, causal, or temporal~\citep{unsworth2014multimodality}).

There is a recent line of work using multivariate information theory to quantify multimodal interactions~\citep{liang2023quantifying}, but there remain exciting new directions extending it to understand pointwise interactions (instead of the distribution-level), for more than 2 modalities, modeling temporal interaction dynamics, for settings requiring sequential prediction reasoning, and causal inference. This line of work can have great impact in developing a more fundamental understanding of the multimodal interactions present in data and those learned by AI models~\citep{liang2023quantifying}. It can also inspire computationally efficient training and inference methods~\citep{han2025guiding}, model architectures~\citep{yu2024mmoe}, and pre-training strategies~\citep{liang2023factorized} that can optimally select and combine the most relevant modalities, alleviating resource requirements and making multimodal technologies more accessible to application domains.

While information theory tells us the type and amount of interactions, it does not specify the ideal fusion function to learn them. Characterizing the types of interactions learned by different architectures, assuming white-box access to the model's architecture~\citep{ittner2021feature}, internal weights~\citep{tsang2018detecting,tsang2019feature}, and fusion functions~\citep{wenderoth2025measuring}, or with only black-box access to input modalities and output representations or labels returned by the multimodal model~\citep{hessel2020emap,liang2023multiviz}, are key challenges that need to be tackled. Other complementary frameworks for different levels and types of synergy should be developed, such as using complexity analysis to understand the complexity of different fusion functions and how they match the interactions in the data, studying what kind of multimodal kernel space permits task separability, and quantifying whether structured or disentangled representations provide provable benefits.

\subsection{Unification vs modularity}

A key scientific question in multimodal learning is the choice between unification versus modularity. There is exciting work showing that models are converging in the representation spaces they learn~\citep{huh2024position}, that modern sequence models are applicable to many modalities~\citep{jaegle2021perceiver,liang2022highmmt,likhosherstov2022polyvit,reed2022generalist}, and that models trained on one modality still encode information about others~\citep{lu2022frozen}. For example, LLMs already have visual priors~\citep{luo2025probing,wang2025words}, robotics and planning capabilities~\citep{huang2022language,li2022pre,wang2024large,zitkovich2023rt}, and social commonsense~\citep{sap2019socialiqa,subramanian2025pose} without ever seeing or experiencing these other modalities. These observations have inspired new directions towards a science of unified multimodal modeling, including architectures that can encode many modalities~\citep{wu2024next,zhan2024anygpt}, scaling laws for unified models (even meta scaling laws that can generalize and predict the scaling behaviors for new modalities)~\citep{deng2025emerging}, and unified data, models, and training to encourage cross-modal transfer~\citep{chen2025blip3,chen2025janus,zhang2025unified}.

At the same time, unified models can struggle with compositionality~\citep{han2024progressive}, lack interpretability~\citep{hewitt2025we}, and are harder to control~\citep{wang2024comprehensive}. These considerations have inspired advances in modularity and compositionality, such as multi-agent LLMs~\citep{du2023improving} and composable modules~\citep{dahlgren2024learning} across modalities. These can improve accuracy especially in complex tasks with long-range and compositional structures. Extending them to more modalities can yield strong advancements. What are the principled theoretical frameworks and practical experiments to study the pros and cons of each paradigm?

\subsection{Learning and optimization}

Multimodal optimization is known to be a core challenge -- as compared to training unimodal models, multimodal models introduce several new challenges. Modality collapse happens when a model defaults to use only a dominant modality and ignore information from non-dominant input modalities~\citep{sim2025can}. Modality interference, competition, or sabotage happens when models use spurious or conflicting information acquired from other modalities, leading to degraded performance~\citep{cai2025diagnosing,huang2022modality,kontras2024multimodal}. Modality imbalance is due to different modalities overfitting and generalizing at different rates due to differing heterogeneity and information, so training them jointly with a single optimization strategy is sub-optimal~\cite{wang2020makes,wu2022characterizing}. Taken together, optimization for multimodal models must be \textit{data-dependent} and explicitly leverage properties of the modalities, their heterogeneity, and interactions, rather than treating optimization as entirely data-agnostic. Early work in data-dependent optimization for unimodal settings~\cite{kunstner2024heavy,lee2025efficient,zhao2024deconstructing} hint at a more general paradigm where optimization, data, and architectures are mutually dependent components of a unified system.

In addition to these challenges, there are also opportunities to develop new training and optimization strategies when multimodal data is present. For example, recent work has shown that directly optimizing for modality synergy, redundancy, uniqueness can provide new training signals not directly captured by supervised prediction~\citep{liang2023quantifying}. Additional modalities or views also inspire new self-supervised training signals including cross-modal prediction~\citep{li2019connecting} or cross-modal contrastive learning~\citep{radford2021learning}. Even in single modality settings, these new objectives have become the backbone for many large-scale pre-trained models, including designing training objectives to capture redundant information across views~\citep{chen2020simple,tian2020makes} (contrastive or non-contrastive), and extending them to capture other interactions~\citep{liang2023factorized}. Finally, we highlight abundant new directions in multimodal optimization, especially those inspired by the brain, such as cross-modal plasticity (re-purposing the functionality of one sensory encoder to process information from another)~\citep{bavelier2002cross}, and cross-modal synesthesia (stimulation of one sensory modality triggering experiences in another)~\citep{marks1975colored}. How can we develop fundamentally new optimization objectives to develop these new multimodal capabilities?

\vspace{-1mm}
\section{The Next Frontier for Multisensory Intelligence}
\label{sec:synergy}
\vspace{-1mm}

To outline a vision for future research in multisensory intelligence, we highlight 6 core technical challenges. These technical challenges build upon, and extends, prior survey papers on multimodal AI~\citep{baltruvsaitis2018multimodal,gan2022vision,jin2024efficient,li2024multimodal,liang2024foundations,xie2024large,xu2023multimodal}. Specifically, the key differences are a larger focus on multimodal foundation models and their pre-training, adaptation, test-time inference, scaling, interaction capabilities, and impact on the human experience.

\subsection{Integration}

\begin{definitionbox}[label=def:integration]{\textbf{Challenge 1, Integration: } }
Learning joint representations that capture cross-modal interactions while accounting for the inherent heterogeneity of different modalities.
\end{definitionbox}

\begin{figure}
    \centering
    \vspace{-4mm}
    \hspace*{-3mm} \includegraphics[width=0.9\linewidth]{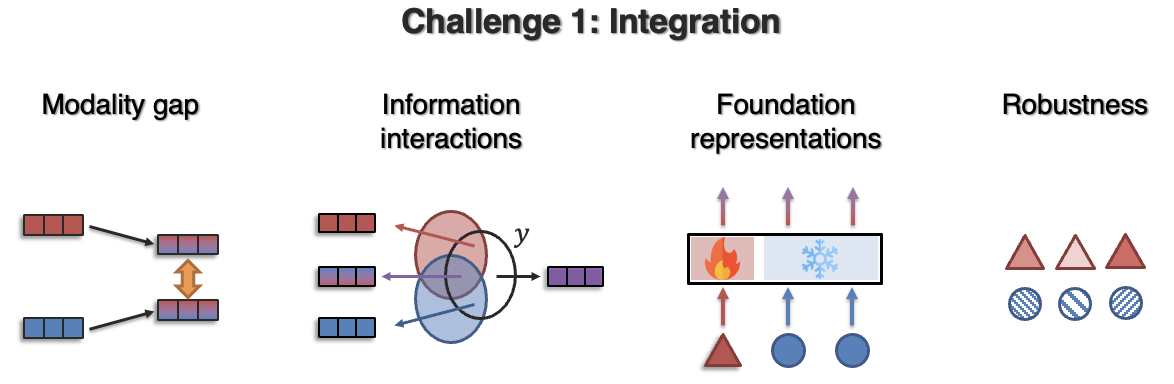}
    \vspace{-1mm}
    \caption{\textbf{Challenge 1, Integration:} Learning joint representations that capture cross-modal interactions while accounting for the inherent heterogeneity of different modalities. Integration requires tackling the challenges of (1) \textit{modality gaps} that cause fundamental representation tensions, (2) \textit{information interactions} that have to be modeled appropriately to maximize task relevance, (3) leveraging and adapting \textit{foundation representations} from pre-trained models, and (4) dealing with \textit{real-world constraints} so that models can work in noisy, imperfect, and resource constrained environments.}
    \label{fig:challenge1}
\end{figure}

Integration is a core building block of multisensory AI, concerned with representing individual sensory elements and their interactions to learn higher-order, informative features. It has a long history in the study of human multisensory integration and efforts to endow machines with similar capabilities~\citep{stein1990multisensory}. Integration is challenging for several reasons. Firstly, the existence of \textit{modality gaps} causes fundamental representation tensions: greater heterogeneity across modalities necessitates more specialized encoders, resulting in increasingly misaligned representations. Secondly, how to learn \textit{information interactions}. As interactions are richer, their representations must become increasingly similar to support alignment, comparison, and fusion. Thirdly, the opportunities and challenges of using \textit{foundation representations} from pre-trained models for quick adaptation to downstream tasks. Finally, dealing with \textit{real-world constraints} in representation integration so that models can work in noisy, imperfect, and resource constrained environments.

\textbf{Modality gap} refers to the phenomenon that jointly representing different modalities with different structures and properties is inherently difficult. As a result, one has to consider the heterogeneity in the modalities and their representations. Fusion of static multimodal data or representations is more straightforward~\citep{liang2024foundations}. Fusing multiple sequences, while harder, can still be done with transformers, state space models, and diffusion sequence models~\citep{qiao2024vl,tsai2019multimodal,yang2025mmada}. The challenge increases when fusing data that is highly heterogeneous, such as static and temporal, spatial and temporal, or discrete and continuous, such as text and audio, video, or sensors~\citep{baris2025foundation,mo2023multiiot}. When modalities are highly heterogeneous, a key question is when to bring heterogeneous representations together. Specialized encoders can capture heterogeneity before alignment, while unified encoders can immediately learn a homogeneous representation space~\citep{hu2021unit}. Recent advances have attempted to better bridge these data forms using hybrid models of transformers with continuous diffusion~\citep{bao2023one,peebles2023scalable}, and unified architectures that can learn from both heterogeneous sources of data (e.g., bridging spatial and temporal, discrete and continuous)~\citep{deng2025emerging,liang2022highmmt,qu2025tokenflow,zhang2025unified}.

\textbf{Information interactions.} \ A key challenge is identifying what information is most desirable to encode in the representation. Through the lens of multimodal interconnections, it can be useful to design representations appropriately for redundant, unique, and synergistic interactions. For example, exploiting multi-view redundancy~\citep{tian2020makes,tosh2021contrastive} is commonly done as a training signal for self-supervised learning. Specific training objectives can include contrastive learning~\citep{radford2021learning}, cross-modal alignment~\citep{rasenberg2020alignment}, and cross-modal prediction~\citep{li2019connecting}. Redundant representations are also beneficial for robustness~\citep{chen2025redundancy,nguyen2025robult,pham2019found}, but lose out on efficiency~\citep{liang2021multibench}. To learn unique representations, modality selection is often performed as a way to identify which modalities to select while minimizing cost and interference from others without useful information~\citep{peng2005feature}. Better quantification methods for modality value~\citep{chen2024quantifying,liang2023quantifying} have been proposed to learn unique representations. Finally, learning representations that provide new synergistic information by learning richer feature spaces~\citep{dong2023dreamllm,liang2023quantifying}, capturing more informative interactions between modalities~\citep{gadzicki2020early}, and contextualizing multiple representations~\citep{tsai2019multimodal}, have been shown to improve performance on downstream tasks requiring synergy.

Key types of modular fusion methods can be applied at different spatial and temporal scales: on top of abstract modalities or unimodal predictions, additive fusion (also known as late fusion)~\citep{baltruvsaitis2018multimodal,friedman2008predictive}, multiplicative interactions~\cite{baron1986moderator,jayakumar2020multiplicative}, and tensor products~\citep{hou2019deep,liu2018efficient,zadeh2017tensor}. Multimodal gated units and attention units learn representations that dynamically change for every input~\cite{arevalo2017gated,chaplot2017gated,wang2020makes}. Fusion today includes late adapter fusion, where pre-trained models and representations are fused via lightweight adapter layers~\citep{gao2023llama,zhang2025videollama}. On the other extreme, early fusion of raw modalities~\citep{baltruvsaitis2018multimodal} can be more challenging since raw modalities are likely to exhibit more dimensions of heterogeneity, but benefits from fine-grained expressivity~\citep{liang2022highmmt,likhosherstov2022polyvit} and increased robustness~\citep{barnum2020benefits,gadzicki2020early}. Recently, early fusion is performed using native pre-training of multiple temporal streams~\citep{hu2021unit,xie2024show,shukor2025scaling,team2024chameleon}. There is also substantial interest in using mixture of experts fusion, multiple LLMs, and multi-agent fusion, where each processor carries their own expertise and are combined through multiple-steps of fusion and reasoning~\citep{ghafarollahi2025automating,li2025uni,yu2024mmoe}.

\textbf{Foundation representations} leverage and adapt large-scale multimodal foundation models as priors for sample efficient inference on downstream tasks. Multimodal foundation models are trending towards unified understanding and generation, using architectures like multimodal transformers~\citep{hu2021unit,liang2022highmmt,xie2024show}, diffusion sequence models and flow-matching~\citep{reuss2024multimodal,yang2025mmada}, and nowadays unified early fusion models~\citep{shukor2025scaling,team2024chameleon}. To adapt these models, key methods include supervised fine-tuning~\citep{liu2023visual}, reinforcement learning and test-time reasoning~\citep{ouyang2022training}, using (often low-rank) adapter layers in the prompt space~\citep{lester2021power,li2021prefix,tsimpoukelli2021multimodal} and representation space~\citep{hu2022lora,ziegler2019encoder}. While foundation model representations are powerful, there remain open questions regarding the next paradigm of foundation models natively pre-trained on video and audio, and how to quickly adapt it for downstream tasks. Furthermore, current methods to adapt foundation representations still employ simplistic adapter or late fusion strategies, such as processing vision and language through separate foundation models before adapting using linear transformations~\cite{bai2025qwen2,liu2023visual}, which may fail to capture rich synergy at spatio-temporal levels~\citep{liang2023quantifying}, compositionality between modalities~\cite{girdhar2023imagebind,thrush2022winoground}, and fine-grained visual details~\citep{tang2025chartmuseum} beyond high-level semantic understanding~\citep{fu2025hidden}. Mitigating undesired cross-modal information loss in simple fusion strategies is a key question in training and adapting multimodal foundation representations.

\textbf{Real-world constraints.} \ Finally, how can we learn multimodal representations and integrate information in the the presence of real-world imperfections, computational constraints, and human-centered safety considerations? It is critical to learn robust features in the presence of missing, noisy, or adversarial modalities~\cite{ding2021multimodal,foltyn2021towards}. There are methods directly developed to handle imperfect data by introducing imperfect data during training~\citep{lee2023multimodal,liang2021multibench,ma2022multimodal}, or post-hoc tuning models to handle imperfect data~\citep{ma2021smil,wang2020multimodal}. A new direction is to train general models with a wide range of modalities so they can still operate on partial subsets (i.e., modality dropout), thereby anticipating missing modalities in advance~\citep{liang2022highmmt,reed2022generalist}.
There also needs more principled quantization and compression techniques for multimodal networks that are increasing in size~\citep{jin2024efficient}. Quantization methods have largely been developed for vision or language models in isolation, even though multimodal systems pose unique challenges, such as aligning precision across heterogeneous feature spaces, maintaining synchronization across sensory channels, and dealing with asymmetric signal-to-noise ratios~\citep{kunstner2024heavy}. More research should also identify how real-world concerns change in the face of different multimodal interactions (e.g., dealing with missing modalities when information is redundant is easier than when there is synergy). Finally, true real-world robustness needs a rethinking of benchmarks from single datapoint test cases to invariants, guarantees, and verifications over all families of multimodal inputs and outputs.

\begin{definitionbox}[label=open:integration]{\textbf{Open directions in integration: } }
\begin{enumerate}[leftmargin=0.5cm,parsep=0pt,partopsep=0pt]
    \item A rigorous science of multimodal information, interactions, and integrated representations.
    \item Models that effectively bridge the modality gap, such as between static and temporal, spatial and temporal, and discrete and continuous data.
    \item Methods to fine-tune and adapt foundation representations for downstream tasks while accounting for verying modality heterogeneity and interactions.
    \item Native multimodal models that learn synergistic interactions at fine-grained spatial and temporal resolutions.
    \item Balancing representation learning with real-world constraints like robustness, safety, and efficiency.
\end{enumerate}
\end{definitionbox}

\subsection{Alignment}

\begin{definitionbox}[label=def:alignment]{\textbf{Challenge 2, Alignment: } }
Modeling fine-grained cross-modal relationships among all modality elements by leveraging the structural dependencies present in the data.
\end{definitionbox}

\begin{figure}
    \centering
    \vspace{-4mm}
    \hspace*{8mm} \includegraphics[width=0.9\linewidth]{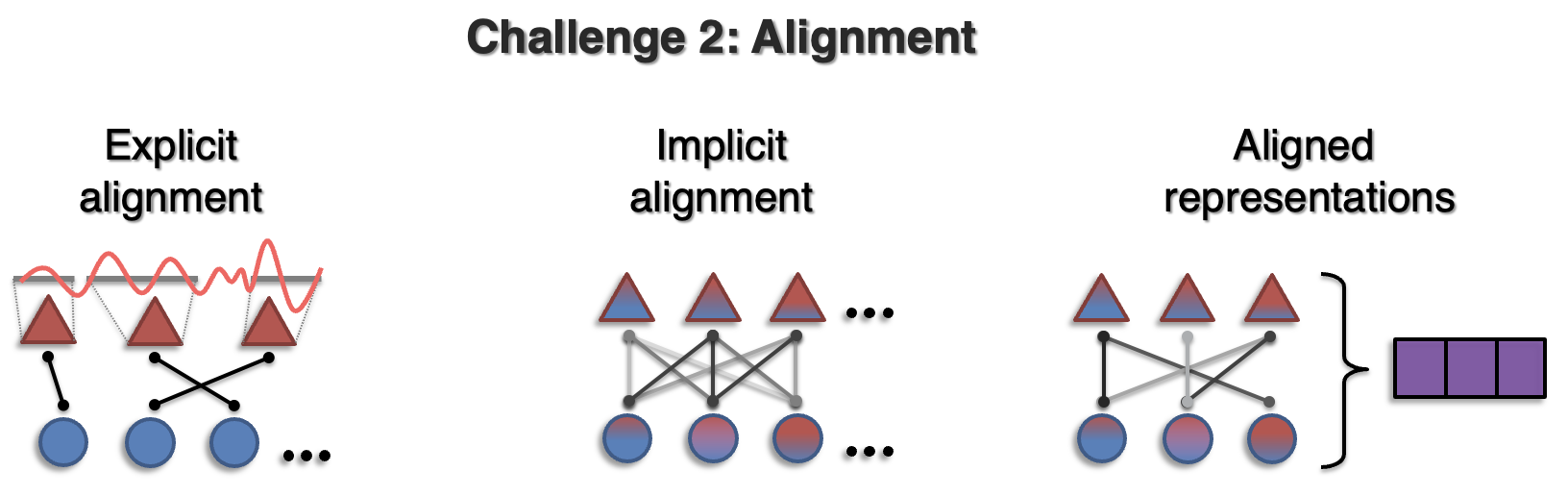}
    \vspace{-1mm}
    \caption{\textbf{Challenge 2, Alignment:} Modeling fine-grained cross-modal relationships among all modality elements by leveraging the structural dependencies present in the data. Alignment contains 3 key subchallenges of (1) \textit{explicit alignment} using learning algorithms to identify connections between discrete elements or continuous modality signals with ambiguous segmentation, (2) \textit{implicit alignment} where alignment is not directly imposed but rather an emergent phenomenon from downstream learning objectives, and (3) \textit{aligned representations} where alignment is a latent step to learn better contextualized multimodal representations. }
    \label{fig:challenge2}
\end{figure}

Alignment seeks to identify cross-modal connections between elements of multiple modalities. For example, when analyzing the speech and gestures of a speaker, how can we align specific gestures with spoken words or utterances?
Alignment between modalities is challenging since it can be multifaceted (one-to-one, many-to-many, or not exist at all), may depend on long-range dependencies, and often involves ambiguous segmentation of continuous data. There are 3 key subchallenges of (1) \textit{explicit alignment} using learning algorithms to directly identify connections between discrete elements or continuous modality signals with ambiguous segmentation, (2) \textit{implicit alignment} where alignment is not directly imposed but rather an emergent phenomenon from downstream learning objectives, and (3) \textit{aligned representations} where alignment is a latent step to learn better contextualized multimodal representations.

\textbf{Explicit alignment.} \ Recent approaches use contrastive learning~\citep{frome2013devise,radford2021learning}, optimal transport~\citep{lee2019hierarchical,pramanick2022multimodal}, and adapter layers~\citep{tsimpoukelli2021multimodal,zhu2023minigpt} to explicit discover connected modalities. Main considerations include deciding on the granularity of aligned data -- whether directly paired at the fine-grained element level (e.g., word token and image region), or at a less granular or high-level pairings (e.g., entire caption and image). Alignment can also be learned from unpaired data using techniques like optimal transport or unsupervised alignment~\citep{liang2024foundations}. Naturally, there are tradeoffs between the ease of obtaining paired data vs difficulty in imposing alignment objectives. Another key consideration is the choice of alignment function. There exist objectives to promote direct one-to-one connections like cosine distance~\cite{mekhaldi2007multimodal} or max-margin losses~\cite{hu2019deep}. Other alignment functions are more suitable to learn many-to-many connections~\citep{alviar2020multimodal}, correlations~\citep{thompson2000canonical}, and their non-linear generalizations~\citep{andrew2013deep,lai2000kernel,rasiwasia2010new,wang2015deep}. Alignment can also be defined over pairwise orderings~\cite{vendrov2015order} and hierarchies~\citep{delaherche2010multimodal,zhang2016learning} across modalities. Finally, theory has shown that contrastive approaches provably capture redundant information across the two modalities~\citep{tian2020contrastive,tosh2021contrastive}, but not non-redundant information.

Finally, another important consideration is how heterogeneity influences alignment. It is much harder to align discrete vs continuous data and spatial vs temporal data, due to differences in their inherent connections. For continuous data such for images~\citep{ramesh2021zero,van2017neural}, video~\citep{sun2019videobert}, and audio~\citep{liu2022cross}, there often needs additional steps of segmentation and tokenization. There has been a recent trend to develop methods that do not require tokenization, which can save on information loss when converting continuous data into discrete boundaries\citep{fan2025fluid,hwang2025dynamic}. If such methods generalize to more modalities, they could be used to directly align data in input space rather than relying on lossy discrete tokens. Alternatively, there also exist promising approaches to bridge heterogeneous modalities through alignment at the tokenizer level, such as text-aligned visual tokenization~\citep{zhao2025qlip}. Other approaches developing unified tokenizers have also achieved strong results on unified multimodal understanding~\citep{qu2025tokenflow,wang2025mio}. These advances pose key questions regarding the level of alignment that should be done in multimodal models, from the raw data and tokenizer level, to early fusion and representation level, or post-hoc alignment (i.e., late fusion) of pre-trained modality encoders.

\textbf{Implicit alignment.} \ In contrast to explicit alignment, implicit alignment studies how alignment can emerge without directly training with an explicit objective function. Several approaches for alignment to emerge include information sharing, either through the same architecture~\citep{chen2025blip3}, parameters~\citep{zhang2025unified}, data formats (interleaved image-text training)~\citep{laurenccon2023obelics}, and naturally scaling data and model parameters~\citep{huh2024position}. Implicitly aligned representations enable alignment to be captured through simple linear transformations; with recent work showing that a single linear mapping can align separately trained vision and language models~\citep{merullo2023linearly}. Crucially, implicitly aligned representations also enable better contextualization for downstream tasks, forming the basics of transformer and other sequence models~\citep{shukor2024implicit}. However, there is a limit to emergent alignment when there is information asymmetry and heterogeneous modalities~\citep{tjandrasuwita2025understanding}. In these settings, alternative approaches extending contrastive learning have been proposed to capture alignment~\citep{liang2023factorized,wang2022rethinking}.

\textbf{Aligned representation} learning aims to model multimodal alignment between connected elements to learn better representations. In other words, alignment is only a \textit{latent} intermediate step enabling better performance on downstream multimodal tasks. This latent alignment can be modeled in an undirected way, where the connections are symmetric in either direction~\cite{li2019visualbert,sun2019videobert}, or via cross-modal directed alignment, where alignment is directed from a source modality to a target modality~\cite{gan2022vision,liang2022highmmt,reed2022generalist,tsai2019multimodal}. Aligned representations can be learned via end-to-end training with multimodal transformers~\citep{alayrac2022flamingo,awadalla2023openflamingo,xu2023multimodal}, or keeping unimodal parts frozen and only training a post-hoc alignment layer~\cite{dai2023instructblip,gao2023llama,li2023blip,zhu2023minigpt}. Self-supervised pretraining has emerged as an effective way to train these architectures to learn aligned representations from larger-scale unlabeled multimodal data before transferring to specific downstream tasks via supervised fine-tuning~\cite{driess2023palm,li2019visualbert,zhu2023minigpt}. Pretraining objectives typically consist of image-to-text or text-to-image alignment~\cite{hendricks2021decoupling,zhu2023minigpt}, and multimodal instruction tuning~\cite{dai2023instructblip,liu2023visual,lu2023empirical}. We refer the reader to recent survey papers discussing these large vision-language models in more detail~\cite{du2022survey,gan2022vision}.

\begin{definitionbox}[label=open:alignment]{\textbf{Open directions in alignment: } }
\begin{enumerate}[leftmargin=0.5cm,parsep=0pt,partopsep=0pt]
    \item Principled ways to decide when and how to align and fuse multimodal representations, and the interplay between alignment and fusion.
    \item Rigorously studying the phenomenon of emergent alignment and its opportunities for representation learning.
    \item Understanding the limits of alignment with respect to modality heterogeneity and interconnections.
    \item Aligned representations that maximize multimodal synergy and integration at fine-grained levels.
    \item Alignment methods for high-frequency and continuous data, without requiring tokenization or segmentation.
\end{enumerate}
\end{definitionbox}

\subsection{Reasoning}

\begin{definitionbox}[label=def:reasoning]{\textbf{Challenge 3, Reasoning: } }
Performing multi-step inference across modalities to integrate and synthesize knowledge guided by task structure.
\end{definitionbox}

\begin{figure}
    \centering
    \vspace{-4mm}
    \hspace*{8mm} \includegraphics[width=0.8\linewidth]{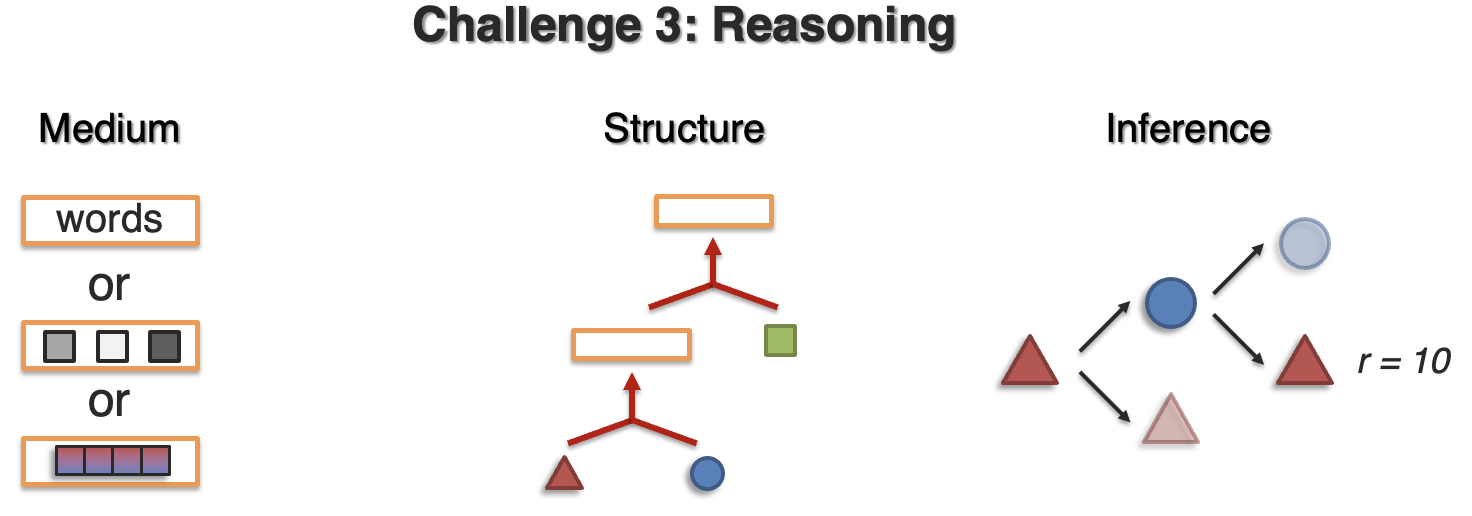}
    \vspace{-1mm}
    \caption{\textbf{Challenge 3, Reasoning:} Performing multi-step inference across modalities to integrate and synthesize knowledge guided by task structure. General multimodal reasoning requires developing (1) \textit{reasoning mediums} that parameterize individual concepts in the reasoning process, (2) \textit{structure modeling} of the relationships over which reasoning occurs, and (3) \textit{inference} of increasingly rich concepts from individual steps of evidence.}
    \vspace{-2mm}
    \label{fig:challenge3}
\end{figure}

Today's language models are increasingly capable of reasoning over multiple steps with interpretable explanations, answer verification, and backtracking to solve challenging language problems~\citep{guo2025deepseek}. However, multimodal reasoning models that can reason over an integrated set of modalities, such as text, images, audio, video, and knowledge graphs, are sorely lacking and can pave the way for the next frontier of AI. To achieve general multimodal reasoning, we identify several subchallenges: (1) \textit{reasoning medium} studies the parameterization of individual concepts in the reasoning process, (2) \textit{structure modeling} involves defining or structuring the relationships over which reasoning occurs, and (3) \textit{inference paradigm} defines how increasingly rich concepts are inferred from individual steps of evidence.

\textbf{Reasoning medium, inputs, and outputs} categorize the intermediate representations used for step-by-step reasoning. Language is the most natural medium for reasoning since it is interpretable by users, compositional in its structure, and easily decomposable into multiple steps~\citep{zeng2022socratic}. This has catalyzed a significant body of work on reasoning in LLMs. At the same time, other mediums have also been used for reasoning. Evidence has shown that reasoning in the latent space of LLMs (rather than raw text tokens) bring practical~\citep{hao2024training} and theoretical benefits~\citep{zhu2025emergence}. Recent work has also explored visual~\citep{shao2024visual,wang2025multimodal} and video~\citep{fei2024video} chain of thought reasoning to solve complex problems involving spatial reasoning, such as in math~\citep{lee2025interactive}, science~\citep{zhang2023multimodal}, and visual data analysis~\citep{kaur2025chartagent}. Attention maps are also a popular choice for visual concepts since they focus reasoning on a certain region to improve human interpretability~\citep{andreas2016neural,xu2015show,zhang2020multimodal}. Finally, we highlight emerging directions in neuro-symbolic mediums that integrate discrete symbols as intermediate reasoning steps~\cite{kambhampati2024position,yang2023neuro}, combining these systems with logic-based differentiable reasoning~\cite{amizadeh2020neuro,cheng2025empowering}, using world models as priors for visual reasoning~\citep{guan2023leveraging,wiedemer2025video}, and multi-lingual reasoning to promote accessibility and inclusion especially for culture-specific tasks~\citep{aggarwal2025language,hwang2025learn}.

Given many possible mediums for reasoning, it becomes critical to study the tradeoffs in reasoning across mediums. Certain mediums like language are more easily decomposable into multiple steps while others like vision have more information per step. Developing optimal and adaptive reasoning procedures will be critical since different problems will have different optimal mediums, or combinations of mediums, for reasoning. Finally, attention architectures and reasoning interplay in higher-order effects -- new alternatives that simulate richer computational capacity or denser connectivity~\citep{roy2025fast,sanford2023representational} and diffusion-based architectures that provide a reasoning ``scratch pad for free''~\cite{ye2025beyond} can potentially enhance the self-correction and multimodal reasoning capabilities of today's models.

\textbf{Reasoning structure} defines how a problem is decomposed, organized, and solved through substeps, whether through linear chains of thought~\citep{kojima2022large, wei2022chain}, tree-structured exploration~\citep{yao2023tree}, or graph-based search~\citep{besta2024graph}. Reasoning must also be adaptive, allowing models to dynamically adjust the number and type of steps taken to balance performance and efficiency~\citep{aggarwal2025l1,zhou2025mem1}. In interactive settings, the structure of reasoning becomes stateful: the agent's internal state, intermediate beliefs, and available actions evolve as it interacts with the environment~\citep{bisk2020experience,luketina2019survey}. Reinforcement learning is frequently used to train such structured decision processes~\citep{chaplot2017gated,narasimhan2018grounding,wang2019reinforced}. In many problems, the reasoning structure is known or can be annotated by domain experts. When the structure is not known, it can instead be learned jointly with the reasoning process, enabling models to optimize latent structures that best support the task~\citep{li2025llms,perez2019mfas,xu2021mufasa}.

Another critical frontier is long-horizon reasoning, particularly as we expand to long-range text, audio, video, and continuous sensing modalities. Token-efficient architectures using linear attention~\citep{li2024videomamba,ren2025vamba,song2025videonsa} and intelligent retrieval systems~\citep{zhang2025deep} promise sustained observation and reasoning over extended time spans. However, existing models still lack inductive biases and positional awareness inherent in older recurrent architectures~\citep{chang2025language}, often requiring far more data to generalize to longer, more compositional tasks. Long-range problems may require revisiting ideas from fixed-memory recurrent paradigms or developing new ones. New efficient reasoning models would enable learning from audio, video, and real-world sensing data, where there are orders of magnitude richer data than internet-scale text.

\textbf{Reasoning training and inference} studies how we can incentivize reasoning of increasingly rich concepts from individual steps of evidence. This can be achieved through explicit training with dense supervision~\citep{ouyang2022training} or RL for step-by-step reasoning to emerge~\citep{shao2024deepseekmath}. Test-time reasoning with RL is a popular paradigm but requires a good reward function or verifier~\citep{rafailov2023direct}. While easier to define for objective language tasks like math and code, there are challenges to define it for subjective human-centered social tasks~\citep{ong2025human}. What rewards should be used to encourage multimodal reasoning, especially cross-modal rewards or multimodal alignment rewards? Another key challenge is balancing heterogeneous data distributions during RL training; classic deep-RL approaches such as task-wise normalization in IMPALA~\citep{espeholt2018impala} and PopArt~\citep{hessel2019multi} explored adaptive rescaling, and recent work has extended these ideas to RL post training of multimodal large language models~\citep{dai2025qoq}.
Integrating more advanced search, sampling, and inference techniques like probabilistic, logical, and counterfactual inference~\citep{agarwal2020towards,amizadeh2020neuro,niu2021counterfactual} could further improve the accuracy and robustness of multimodal reasoning. Finally, an important future direction could lie in using diffusion sequence models for multimodal reasoning, enabling latent reasoning in multiple modalities and across continuous steps.

\begin{definitionbox}[label=open:reasoning]{\textbf{Open directions in reasoning: } }
\begin{enumerate}[leftmargin=0.5cm,parsep=0pt,partopsep=0pt]
    \item Multimodal reasoning across input and output mediums such as text, images, video, symbolic structures, different languages, and their combinations.
    \item Studying the tradeoffs between reasoning in different mediums for optimal, efficient, and adaptive reasoning, especially over long-horizon modalities.
    \item Reasoning over structures (e.g., chains, trees, graphs, multi-agent) to reflect dependencies across modalities.
    \item Exploring new ways to incentivize reasoning when there are ambiguous rewards or limited supervision.
    \item Combining reasoning with expressive causal, counterfactual, and compositional inference methods.
\end{enumerate}
\end{definitionbox}

\subsection{Generation}

\begin{definitionbox}[label=def:generation]{\textbf{Challenge 4, Generation: } }
Learning a generative process that synthesizes raw multimodal data by modeling cross-modal interactions, modality structure, and maintaining coherence across modalities.
\end{definitionbox}

\begin{figure}
    \centering
    \vspace{-6mm}
    \hspace*{-3mm} \includegraphics[width=0.73\linewidth]{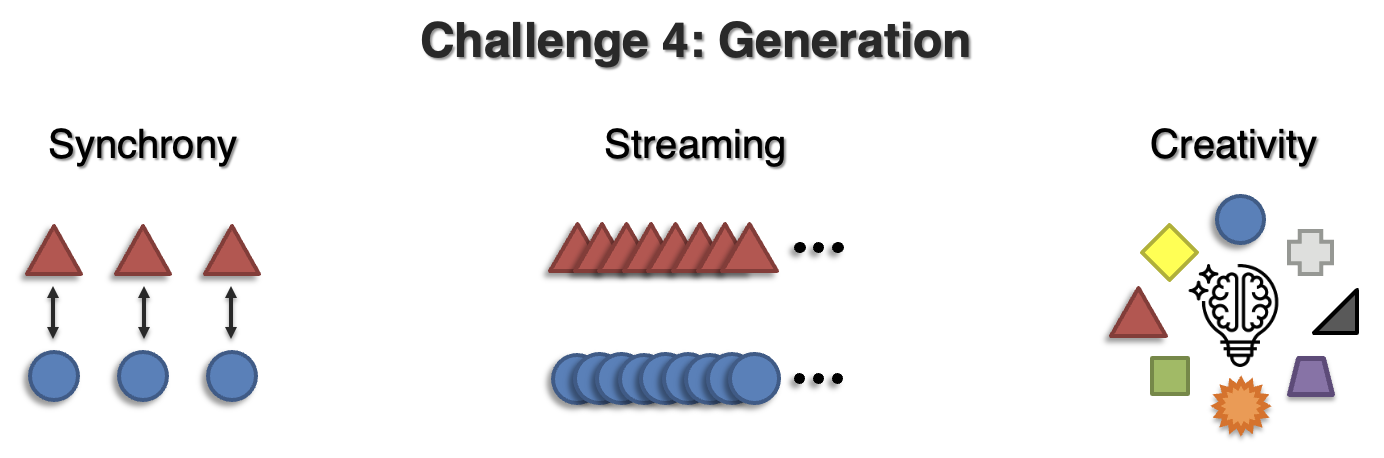}
    \vspace{-2mm}
    \caption{\textbf{Challenge 4, Generation:} Learning a generative process that synthesizes raw multimodal data by modeling cross-modal interactions, modality structure, and maintaining coherence across modalities. Modern multimodal generation requires tackling the challenges of (1) \textit{synchrony}: ensuring temporal and semantic alignment across modalities during generation, (2) \textit{streaming}: enabling real-time, continuous multimodal generation that adapts to dynamic inputs, and (3) \textit{creativity}: producing novel cross-modal outputs that inspire and assist users beyond direct imitation.}
    \label{fig:challenge4}
\end{figure}

Generation has been at the forefront of AI research with substantial innovations in language, image, video, and audio generation. The multimodal side has focused on cross-modal generation, such as text to image, video, and audio, image to video, or adding other conditional controls, including stylistic attributes and personas~\citep{cao2025survey}. There is also a trend towards unified multimodal generation across language, video, and audio~\citep{qu2025tokenflow,zhang2025unified} for human interaction, entertainment, and world modeling. Several key challenges in generation include \textit{synchrony}, \textit{streaming}, and \textit{creativity}.

\textbf{Synchrony} is a key quality to ensure temporal and semantic alignment across generated modalities. We break down synchrony into three main types: within-modality coherence, cross-modal coherence, and multimodal coherence. 
Within-modality coherence focuses on maintaining internal structural consistency within each modality, such as spatial coherence in images~\citep{ho2020denoising}, semantic and discourse coherence in text~\citep{brown2020language}, and spatio-temporal coherence in video~\citep{ho2022video}, and is largely addressed by state-of-the-art unimodal generative models.
Cross-modal coherence refers to conditioning on appropriate control modalities and variables to generate semantically meaningful outputs in another modality. Common examples include faithful text-to-image generation and image-to-text captioning grounded in visual content. Control modalities may include text descriptions~\citep{ramesh2021zero}, target styles~\citep{zhang2023adding}, images or videos~\citep{wang2018video}, as well as structured signals such as robot goals or action trajectories~\citep{reed2022generalist,rt22023arxiv}. Key approaches for achieving cross-modal control include collecting paired multimodal data and learning aligned cross-modal representations~\citep{radford2021learning}, adapting large pre-trained generative models through parameter-efficient finetuning~\citep{zhang2023adding}, and using classifier-based guidance~\citep{dhariwal2021diffusion} or classifier-free guidance~\citep{saharia2022photorealistic} to condition pre-trained generative models. Finally, multimodal coherence aims to preserve consistency across multiple generated modality streams in parallel, such as speech and lip movements or visual content and audio signals. Recent approaches use end-to-end multimodal joint training~\citep{bai2025qwen2,cui2025emu35nativemultimodalmodels} or post-hoc alignment or refinement strategies that enforce cross-modal consistency after independent generation~\citep{comunita2024syncfusion}.

\textbf{Streaming.} \ Most real-world interactive systems require streaming multimodal generation, such as a live streaming avatar, an interactive game, and real-time sports commentary. This poses significant challenges as today's models often generate fixed-length sequences (e.g., a video at once)~\citep{hongcogvideo,wan2025wanopenadvancedlargescale} or perform post-hoc generation (e.g., captioning only after the full video is seen)~\citep{zhang2023video}. Achieving streaming generation requires causal attention. Recent methods such as diffusion forcing~\citep{chen2024diffusion} and self forcing~\citep{huang2025self} enable stable, causal generation for models originally trained with bidirectional attention or full-sequence generation objectives. Temporally synchronized data is often collected to fine-tune or post-train models for valid streaming content generation~\citep{chen2025livecc}. Efficiency methods include model distillation~\citep{yin2024one}, attention manipulation (e.g., sliding-window attention~\citep{bai2025qwen2}), and KV caching techniques~\citep{yang2025longliverealtimeinteractivelong}. For long-horizon generation, long-term memory can be further improved through memory compression methods, such as latent tokens that retain spatial or temporal context~\citep{wu2025videoworldmodelslongterm}.

\textbf{Creativity.} Artistic creation is a multi-stage process, often involving multiple references across mediums and time spans, as well as critique and feedback from different audiences. While AI has demonstrated strong capabilities to generate realistic and high-quality content across many modalities, their potential to support and advance the creative and artistic workflow of creators remains unexplored~\citep{chung2022artistic,doshi2024generative,epstein2023art,zhou2024generative}. We envision several key directions in multimodal and generative AI that need to be tackled. First is to increase the diversity of generated content so that they represent various regions and demographics, and are able to accurately assist in different user contexts~\citep{cai2023designaid,liang2021towards}. Secondly, these systems should enable users the agency to create and choose desired outputs rather than generating outputs completely~\citep{shi2023understanding,wu2021ai}. They also need to be explainable, by explaining how they are generating their current output alongside referencing closest datapoints from their training data~\citep{wang2023evaluating} for proper copyright and attribution. Finally, an important aspect of creativity involves productive struggle; systems should use scaffolding techniques~\citep{gibbons2002scaffolding} where temporary support is provided to help users complete tasks beyond their current unaided ability, but gradually removed to improve their capabilities, creativity, and independence over time~\citep{boussioux2024crowdless,eapen2023generative,rafner2023creativity}.

\begin{definitionbox}[label=open:generation]{\textbf{Open directions in generation: } }
\begin{enumerate}[leftmargin=0.5cm,parsep=0pt,partopsep=0pt]
    \item Synchronizing unimodal generative models with limited multimodal data for multimodal generation.
    \item Unified multimodal understanding and generation within a shared representation space.
    \item Efficient multimodal generative models capable of real-time streaming generation over long-horizons and high-modality settings.
    \item Generative models that support interactive conditioning on multimodal inputs, actions, and feedback.
    \item Generating multimodal content that adapts to human intent, context, and preferences while preserving human-AI co-creativity and agency.
\end{enumerate}
\end{definitionbox}

\subsection{Generalization}

\begin{definitionbox}[label=def:generalization]{\textbf{Challenge 5, Generalization: } }
Predictably understanding knowledge transfer from high-resource modalities or pre-trained models to low-resource ones, including potential performance gains and associated risks.
\end{definitionbox}

\begin{figure}
    \centering
    \vspace{-6mm}
    \hspace*{17mm} \includegraphics[width=0.87\linewidth]{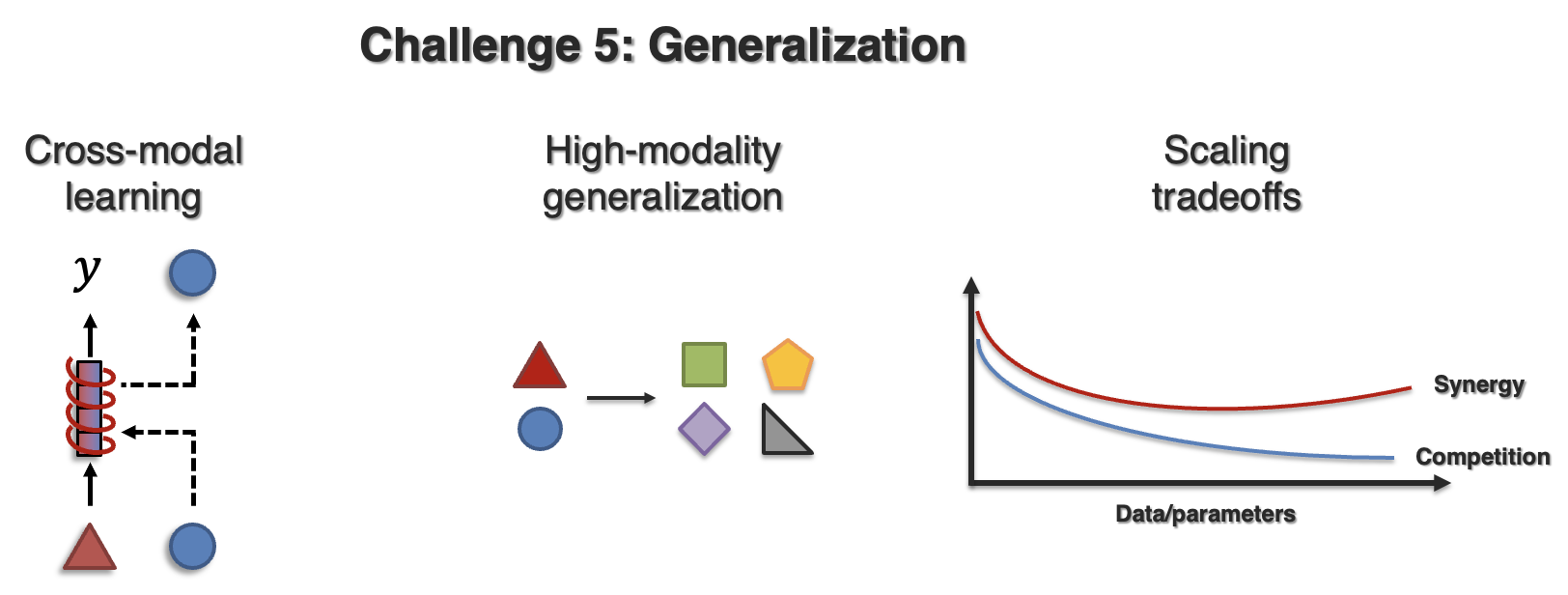}
    \vspace{-2mm}
    \caption{\textbf{Challenge 5, Generalization:} Predictably understanding knowledge transfer from high-resource modalities or pre-trained models to low-resource ones, including potential performance gains and associated risks. We identify three types of generalization: (1) \textit{cross-modal transfer} from high-resource modalities or pre-trained models to low-resource ones, (2) \textit{high-modality generalization} where transfer happens between many modalities, each with only partially observable subsets of data, and (3) \textit{scaling tradeoffs} of how multimodal capabilities and risks change with increased data, model, and training scale to best utilize constrained resources.}
    \label{fig:challenge5}
\end{figure}

Generalization is key to intelligence. Multimodal models often exploit cross-modal generalization via pretrained foundation models, supporting zero-shot, few-shot, and in-context learning, as well as test-time adaptation. It is particularly important to study three types of generalization: (1) \textit{cross-modal transfer} from high-resource modalities or pre-trained models to low-resource ones, (2) \textit{high-modality generalization} where transfer happens between many modalities, each with only partially observable subsets of data, and (3) \textit{scaling tradeoffs} -- predictably quantifying how multimodal capabilities and risks change as a function of increased data, model, and training scale in order to best utilize constrained resources.

\textbf{Cross-modal transfer} seeks to transfer knowledge or representations learned in one modality to improve learning or performance in another modality, enabling more efficient and integrated multimodal understanding. There are several settings when cross-modal transfer helps, including but not limited to (1) when there is redundancy and a second modality reinforces the prediction~\citep{zadeh2020foundations}, (2) when there is uniqueness and information is provided about new concepts or relationships between concepts~\citep{liang2023factorized}, and (3) when there is synergy and new information arises from cross-modal transfer~\citep{aghajanyan2023scaling}. At a high level, there are three ways of encouraging transfer, through (1) cross-modal prediction~\citep{li2019connecting}, (2) alignment~\citep{radford2021learning}, and (3) fusion~\citep{zadeh2020foundations}. Cross-modal prediction learns a translation model from the primary to secondary modality, resulting in enriched representations of the primary modality that can predict both the label and `hallucinate' secondary modalities. Examples include language modeling by mapping contextualized text embeddings into images~\citep{tan2020vokenization}, image classification by projecting image embeddings into word embeddings~\citep{socher2013zero}, and language sentiment analysis by translating language into the speaker's visual and vocal expressions~\citep{pham2019found}. Cross-modal prediction has also been widely applied for robotics~\citep{li2019connecting} and medical applications~\citep{yang2022artificial}. Cross-modal alignment learns an aligned representation space by integrating multiple modalities during training to enable effective knowledge transfer to a primary modality at test time. CLIP, through large-scale image-text alignment training~\citep{radford2021learning}, yields strong vision and language encoders that often outperform vision-only and text-only encoders. Recent work has extended cross-modal alignment to other modalities~\citep{liu2025calf,shen2023cross}. Finally, cross-modal fusion is possible without explicit objectives and training only for fusion~\citep{zadeh2020foundations}. By training models directly on many modalities through autoregressive or masked prediction~\citep{zhang2025unified}, we can learn representations that display transfer, without needing to directly perform cross-modal prediction or alignment which can introduce new optimization challenges. Performing this at scale, and predictably understanding when cross-modal transfer emerges, are key questions that can influence modeling.

In contrast to cross-modal learning, alternative approaches maintain pretrained unimodal models separately and integrate them by exchanging prediction information. For example, co-training uses each unimodal model's predictions to pseudo-label new unlabeled examples in the other modality, thereby enlarging the training set of the other modality~\cite{blum1998combining,cheng2016semi,dunnmon2020cross,hinami2018multimodal}. Co-regularization regularizes the predictions from separate unimodal classifiers to be similar, thereby encouraging both classifiers to share information (i.e., redundancy)~\citep{hsieh2019adaptive,sindhwani2005co,sridharan2008information,yang2019comprehensive}. Applying these methods to ensemble multiple predictions from pre-trained LLMs and other multimodal models can be a promising approach, since they complement well with the prompting and black-box nature of these models.

\textbf{High-modality generalization} refers to transfer between many modalities, each with only partially observable subsets of data. For example, it is common to have paired datasets of image and text, and text and audio, but without image and audio. This problem of partial observability worsens when the number of modalities increases~\citep{liang2022highmmt} and as we go to sensitive domains~\citep{liang2021towards}. Recent work has investigated unsupervised learning from unpaired language and vision~\citep{gupta2025better}. Other approaches use language~\citep{girdhar2023imagebind} and vision~\citep{zeng2022socratic} as pivot modalities for transfer. These extend cross-modal prediction and alignment objectives from only two modalities to a high-modality setting~\citep{han2023imagebind,wang2025omnibind}. Another line of promising work seeks to transfer via large-scale unsupervised training from many modalities (i.e., autoregressive or masked), without explicit objectives -- this can pave a promising future for generalizable multimodal AI~\citep{huang2025open,xia2023achieving}. High-modality generalization will be a key training recipe as we scale foundation models to increasingly many modalities, and coming up with a more principled science is crucial.

\textbf{Scaling tradeoffs.} \ How can we predictably quantify how multimodal capabilities and risks change as a function of increased data, model, and training scale? Given additional data or parameter budgets, these guidelines should help practitioners decide whether to spend it on more unimodal data or multimodal data, and whether to scale unimodal or multimodal components. For example, when do phenomenon like modality synergy, alignment, and cross-modal transfer happen? Do they happen at small scales or only emerge with enough data or large models? Recent work has also attempted to compare the scaling properties of autoregressive~\citep{henighan2020scaling}, contrastive~\citep{cherti2023reproducible}, and native multimodal models~\citep{shukor2025scaling}. In addition to positive scaling, there are observations that parts of multimodal learning get worse with scale, such as increased optimization difficulties~\citep{kontras2024multimodal,wang2020makes} and modality competition~\citep{aghajanyan2023scaling}. Furthermore, we envision a research agenda towards meta scaling laws -- seeking to understand how scaling changes across modalities and tasks, enabling one to predict how a completely new modality would scale~\citep{brandfonbrener2024loss,ruan2024observational}, without having access to much data or models in that modality in the first place.

\begin{definitionbox}[label=open:generalization]{\textbf{Open directions in generalization: } }
\begin{enumerate}[leftmargin=0.5cm,parsep=0pt,partopsep=0pt]
    \item New objectives to promote cross-modal transfer such through large-scale alignment and prediction.
    \item Studying the emergence of cross-modal transfer in multimodal foundation models with respect to data, architectures, learning, and scale.
    \item High-modality generalization across many modalities when only a subset are fully observed.
    \item Principled tradeoffs of scaling, including its effect on modality synergy, competition, transfer, and other emergent capabilities.
    \item Modality and data acquisition frameworks to balance performance and sample efficiency.
\end{enumerate}
\end{definitionbox}

\subsection{Experience}

\begin{definitionbox}[label=def:experience]{\textbf{Challenge 6, Experience: } }
Designing intelligent systems that interact with humans and the environment over multiple steps to enhance productivity, creativity, wellbeing, and human-AI experiences.
\end{definitionbox}

\begin{figure}
    \centering
    \vspace{-6mm}
    \hspace*{-10mm} \includegraphics[width=0.76\linewidth]{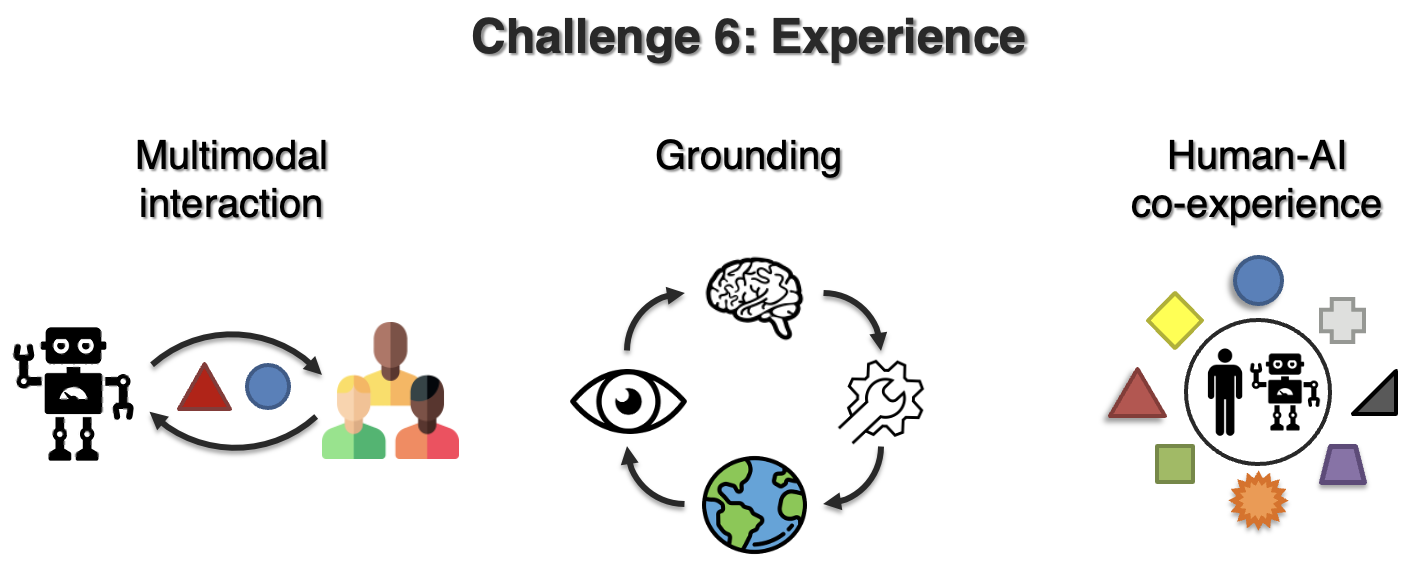}
    \vspace{-2mm}
    \caption{\textbf{Challenge 6, Experience:} Designing intelligent systems that interact with humans and the environment over multiple steps to enhance productivity, creativity, wellbeing, and human-AI experiences. Key subchallenges include (1) developing technologies capable of seamless and adaptive \textit{multimodal interaction}, (2) \textit{grounding} intelligent agents in the social and physical worlds, and (3) fostering \textit{human–AI co-experiences} that enhance productivity, creativity, wellbeing, and human flourishing.}
    \label{fig:challenge6}
\end{figure}

The next wave of AI must be interactive, capable of executing multiple sequences of actions to solve complex, long-horizon problems alongside humans and within physical environments. We envision these technologies augmenting and enhancing human experiences by extending human perception, cognition, and emotional intelligence to enable richer, adaptive, and empathetic interactions. To achieve these goals, key subchallenges include: (1) developing technologies for seamless and adaptive \textit{multimodal interaction}, (2) \textit{grounding} intelligent agents in the social and physical worlds to enable situated and appropriate behavior, and (3) fostering \textit{human–AI co-experiences}: shared interactions between humans and AI systems that enhance productivity, creativity, wellbeing, and human flourishing.

\textbf{Multimodal interaction.} \ How can we define principles and best practices for human-AI interaction across multiple modalities? This question cannot be answered alone from an AI perspective, and needs to synthesize complementary knowledge from AI, HCI, multimodal interaction, psychology, and cognitive science. From an AI perspective, these models need to take in arbitrary modalities as conditioning inputs, generate many modalities via streaming outputs, and maintain memory and coherence over long interaction periods, as illustrated in the previous aims. From a HCI perspective, seminal works by~\citet{norris2004analyzing},~\citet{oviatt1999ten}, and~\citet{turk2014multimodal} have summarized key guidelines for multimodal interaction between people and machines. Many of these guidelines are still relevant today, but many also have to be revamped for modern multimodal generative AI. Research should summarize guidelines for the combinations of interaction mediums (speech, vision, haptics, AR/VR, olfaction), style (formal, playful, collaborative), content (task, narrative, affective), and duration (short feedback vs long-term engagement). Ensuring alignment with FATE principles (fairness, accountability, transparency, ethics) is also central for designing human-centered interactions that are effective, safe, and meaningful. Finally, we envision that the next generation of intelligent multimodal interaction should be aware of ambiguity and uncertainty to be more interpretable for users~\citep{amershi2019guidelines,chen2020investigation,palit2023towards}.

\textbf{Grounded multimodal agents.} \ How can interactive agents model social and physical dynamics within real-world constraints to interact safely and meaningfully with people? We outline several key challanges for developing multimodal agents. First is the ability to ground agent actions in diverse agentic environments -- these include websites, operating systems, databases, engineering pipelines, social interactions, and more. There have been substantial efforts in curating large-scale grounded agent environments~\citep{li2023can,ning2025surveywebagentsnextgenerationai,tbench_2025,zhou2024sotopiainteractiveevaluationsocial}. Secondly, developing approaches for agents to propose diverse actions, verify or reward them appropriately, and efficiently search for optimal long-term action trajectories~\citep{ragen}. Recent advancements in agents decouple high-level planning from low-level action taking~\citep{wang2024voyager}, synergize memory and reasoning for long horizon tasks~\citep{he2025clarabridgingretrievalgeneration,zhou2025mem1}, and enhance agentic models with better search and exploration strategies~\citep{qu2025pope,yang2025treerpotreerelativepolicy}. Self-evolving agents that can automatically propose new tasks and verify progress towards those tasks~\citep{qiu2025alitageneralistagentenabling,simateam2025sima2generalistembodied} are a promising strategy; but perhaps even better are co-evolving agents that self-improve in a way complementary to humans in the loop. Such agents should better communicate uncertainty and defer to human experts for help~\citep{kirchhof2025position}, while learning rapidly from feedback~\citep{shao2025collaborativegymframeworkenabling}, so that intervention is minimized over time~\citep{jin2025erarealworldhumaninteraction}.

Finally, these agents must be grounded in the digital, physical, and social worlds. This entails reasoning about physical laws, social norms, and collective behavior to generate grounded actions that are interpretable, safe, and effective~\citep{hafner2025mastering}. To train these agents, recent trends point towards world models: interactive video generation models that can condition on user actions, prompts, and preferences. However, there remain key challenges for world models to be physically grounded, consistent, and capture multimodal conditioning and feedback. These world models must also be significantly more efficient to enable real-time interactive agents. Extending physical world models to the social and biological worlds can also bring substantial impact: socially-intelligent AI agents will be able to sense, perceive, reason about, learn from, and respond to affect, behavior, and cognition of other agents (human or artificial)~\citep{mathur2025social,zhou2025social}. Biological agents and world models have the potential to transform human health and wellbeing, going from predictive AI trained on passive data~\cite{dai2025qoq,saab2024capabilities,sun2024medical} to active models that combine perception, reasoning, and interaction for active clinical assistance in smart homes, ICU rooms, and surgical rooms~\cite{dai2025developing,intelligence2025pi_,rui2025improving}.

\textbf{Human–AI co-experience.} \ How can multisensory AI create or modify our senses to craft entirely new, immersive, and meaningful experiences? Simply static generation is insufficient, we need to develop models that enhance the joint human–AI co-experience, where jointly constructed, multimodal, and adaptive experiences emerge when humans and AI systems interact, learn, and reason together over time. For example, we envision that future models will generate synchronized visual, auditory, and haptic feedback for gaming and entertainment, create new olfactory and gustatory experiences for personalized diets and health, and design inspiring environments to support creative work and education. These models must sense and respond to human states (attention, emotion, memory) in real time~\citep{picard2000affective} and choose appropriate mediums (text, voice, avatar, embodiment) for interaction~\citep{oviatt1999ten,oviatt2007multimodal}. Other design considerations such as interfaces that adapt dynamically to user attention, cognitive load, and preferences will also be critical~\citep{amershi2019guidelines,shen2024towards,yang2020re}. To evaluate co-experience quality, we need to move beyond single-turn metrics to longitudinal, multi-turn measures (e.g., rapport, affective synchronization, mutual satisfaction) and design benchmarks and data collection protocols that capture the process of co-experience, not just outcomes~\citep{chang2025chatbench}.

\begin{definitionbox}[label=open:experience]{\textbf{Open directions in experience: } }
\begin{enumerate}[leftmargin=0.5cm,parsep=0pt,partopsep=0pt]
    \item Principles of multimodal interaction that capture how humans and AI communicate across sensory channels.
    \item Modeling human intent, attention, and affect for adaptive and empathetic multimodal interaction.
    \item Grounding world models in social and physical commonsense to ensure situated and appropriate interactions.
    \item Using AI to create or modify our senses to craft entirely new, immersive, and meaningful experiences.
    \item Interactive learning and feedback loops that allow humans and AI to co-adapt and improve through shared multisensory experiences.
\end{enumerate}
\end{definitionbox}

\vspace{-1mm}
\section{Conclusion}
\vspace{-1mm}

This vision paper summarizes the key research agenda in multisensory intelligence for the next decade. This multi-disciplinary field extends multimodal AI on the digital language, vision, audio modalities to true intelligence connected to the human senses, social interactions, and the physical world. Multisensory intelligence can significantly enhance human-AI interaction for improved productivity, creativity, and wellbeing. We outlined three interrelated research themes: (1) \textit{sensing} the world through novel modalities and using AI to transform them into structured representations that support learning, reasoning, and decision-making; (2) developing a \textit{science} for the systematic understanding and discovery of generalizable principles from multisensory data; and (3) learning \textit{synergy}, the emergent integration of multiple modalities to create intelligent capabilities greater than the sum of their parts. Together, these research themes pave a way towards synergistic interaction and co-experience between humans and multisensory AI. We encourage community involvement and expansion of these visions at \url{https://mit-mi.github.io/}.

\vspace{-1mm}
\section*{Acknowledgements}
\vspace{-1mm}

We thank all members of the Multisensory Intelligence Group at MIT for helpful discussions, content, references, and feedback; these include (in no particular order) Chanakya Ekbote, David Dai, Megan Tjandrasuwita, Ao Qu, Anku Rani, Ray Song, Lucy Zhao, Jaedong Hwang, Keane Ong, Kaichen Zhou, and Fangneng Zhan.

{\small
\bibliographystyle{plainnat}
\bibliography{refs}
}

\end{document}